\pdfoutput=1
\documentclass[final,3p,times,twocolumn]{elsarticle}

\usepackage{algorithmic}
\usepackage{graphicx}
\usepackage{textcomp}
\usepackage{xcolor}
\usepackage[figuresright]{rotating} 
\usepackage{ragged2e} 
\usepackage{booktabs,makecell, multirow, tabularx}
 
\usepackage{algorithm}
\usepackage{lineno,hyperref}

\usepackage{amsmath}
\usepackage{amssymb}
\usepackage{subcaption}

\journal{Journal of \LaTeX\ Templates}

\begin{document}

\begin{frontmatter}

\title{
    IPENS:Interactive Unsupervised Framework for Rapid Plant Phenotyping Extraction via NeRF-SAM2 Fusion
}
\author[1,2]{Wentao Song}

\affiliation[1]{organization={Hefei Institutes of Physical Science, Chinese Academy of Sciences},
            city={Hefei},
            postcode={230031}, 
            country={China}}

\affiliation[2]{organization={University of Science and Technology of China},
            city={Hefei},
            postcode={230026}, 
            country={China}}
\author[1]{He Huang}

\author[1]{Youqiang Sun}


\author[1,2]{Fang Qu}

\author[1,2]{Jiaqi Zhang}

\affiliation[3]{organization={Anhui Agricultural University},city={Hefei}, postcode={230036}, country={China}}
\author[1,3]{Longhui Fang}

\author[1,2]{Yuwei Hao}

\affiliation[4]{organization={Anhui Jianzhu University},city={Hefei}, postcode={230601}, country={China}}
\author[1,4]{Chenyang Peng}

\begin{abstract}

Advanced plant phenotyping technologies play a crucial role in targeted trait improvement and accelerating intelligent breeding. Due to the species diversity of plants, existing methods heavily rely on large-scale high-precision manually annotated data. 
For self-occluded objects at the grain level, unsupervised methods often prove ineffective.
This study proposes IPENS, an interactive unsupervised multi-target point cloud extraction method. 
The method utilizes radiance field information to lift 2D masks, which are segmented by SAM2 (Segment Anything Model 2), into 3D space for target point cloud extraction. 
A multi-target collaborative optimization strategy is designed to effectively resolve the single-interaction multi-target segmentation challenge.
Experimental validation demonstrates that IPENS achieves a grain-level segmentation accuracy (mIoU) of 63.72\% on a rice dataset, with strong phenotypic estimation capabilities: grain volume prediction yields R² = 0.7697 (RMSE = 0.0025), leaf surface area R² = 0.84 (RMSE = 18.93), and leaf length and width predictions achieve R² = 0.97 and 0.87 (RMSE = 1.49 and 0.21).
On a wheat dataset, IPENS further improves segmentation accuracy to 89.68\% (mIoU), with equally outstanding phenotypic estimation performance: spike volume prediction achieves R² = 0.9956 (RMSE = 0.0055), leaf surface area R² = 1.00 (RMSE = 0.67), and leaf length and width predictions reach R² = 0.99 and 0.92 (RMSE = 0.23 and 0.15).
This method provides a non-invasive, high-quality phenotyping extraction solution for rice and wheat. Without requiring annotated data, it rapidly extracts grain-level point clouds within 3 minutes through simple single-round interactions on images for multiple targets, demonstrating significant potential to accelerate intelligent breeding efficiency.

\end{abstract}


\end{frontmatter}


\section{Introduction}
Plant phenotypic analysis stands as a central direction in modern agricultural research. By elucidating the coordinated expression mechanisms of genomic information and environmental factors \cite{Ying-Hong175}, it provides theoretical foundations for establishing genotype-phenotype association models \cite{genomeassociation}. This research framework holds significant application value for achieving targeted trait improvement in plants and accelerating intelligent breeding processes \cite{ARAUS201452,ARAUS2018451,ZhaoChunjiangCurrentStatus}.
Current phenotyping technologies primarily focus on the visual interpretation of plant morphological features under multi-environment conditions and the dynamic monitoring of physiological traits, involving precise quantification and systematic evaluation of multidimensional phenotypic parameters \cite{AKHTAR2024109033}. Traditional phenotyping methods often rely on manual sampling and invasive detection, facing technical limitations such as limited measurement dimensions, low throughput, and non-reusability of samples, which severely constrain the efficiency of large-scale breeding studies.
Consequently, the development of non-invasive phenotyping extraction technologies based on optical reconstruction and deep learning algorithms, along with the construction of automated phenotyping platforms, has emerged as a cutting-edge research hotspot in the interdisciplinary fields of plant genomics and intelligent agriculture \cite{li2025survey3dreconstructiontechniques}.


Compared to traditional two-dimensional (2D) phenotyping methods \cite{LI2020105672}, three-dimensional (3D) phenotyping technologies offer more comprehensive and accurate representations of plant structures.
These advancements enable researchers to analyze complex morphological features including shape, area, and angles with enhanced precision \cite{LiDARPheno,rs11010015}. 
In recent years, Neural Radiance Fields (NeRF) \cite{mildenhall2021nerf} and 3D Gaussian Splatting (3DGS) \cite{kerbl3Dgaussians} have gained prominence for their ability to generate highly detailed and realistic 3D representations.
Choi et al.\cite{choi2024nerfbased} demonstrated the superiority of 3D phenotyping by automatically collecting tomato RGB images using unmanned greenhouse robots and reconstructing dense point clouds via NeRF. Their results showed significantly improved accuracy compared to traditional 2D approaches. Similarly, PanicleNeRF\cite{YANG20240279} achieved high correlations between 3D point cloud-derived panicle volume and grain traits in rice (indica: R² = 0.85 for grain count, R² = 0.80 for grain weight; japonica: R² = 0.82 and R² = 0.76, respectively) by integrating SAM \cite{kirillov2023segany} for 2D segmentation and NeRF for 3D reconstruction .
Saeed et al.  \cite{10209004}applied NeRF to reconstruct peanut pods from 2D images, achieving a validation set IoU of 0.5 and precision of approximately 0.7, highlighting NeRF’s potential for precise pod localization in legume crops. Shen et al. \cite{SHEN2025110320} combined 3DGS with SAM to reconstruct rapeseed plants using 36 oblique UAV images, enabling accurate biomass estimation. 
Jiang et al.\cite{JIANG2025110293}  utilized 3DGS to extract point clouds of individual cotton plants. They applied YOLOv8x\cite{reis2024realtimeflyingobjectdetection} to generate 2D masks, which were then integrated into  SAGA\cite{cen2023saga} to extract 3D models of cotton bolls and the main stem. This approach enabled the estimation of boll count and main stem length. The results demonstrated mean absolute percentage errors (MAPE) of 11.43\% for boll count and 10.45\% for main stem length, validating the effectiveness of 3DGS in precise, non-destructive plant phenotyping. 
Although 3DGS exhibits higher reconstruction efficiency than lightweight models like Instant-NGP \cite{mueller2022instant}, its performance suffers in complex experimental environments, often resulting in low reconstruction accuracy or poor point cloud quality. In contrast, NeRF’s data-driven nature ensures robust phenotypic trait analysis under diverse field conditions, making it a powerful tool for high-precision plant phenotyping in this study.

Although 3DGS offers higher reconstruction efficiency compared to lightweight models like Instant-NGP\cite{mueller2022instant}, it may encounter issues such as reduced reconstruction accuracy or poor point cloud quality in environments with diverse and complex backgrounds. On the other hand, NeRF adapts to various field conditions, making it a powerful tool for high-precision plant phenotyping.

Accurate 3D object point cloud segmentation constitutes an essential component in the extraction and analysis of 3D phenotypic data.
Deep learning approaches have demonstrated great potential in the field of plant organ segmentation. Supervised methods, after being trained on large volumes of annotated data, can achieve high segmentation accuracy of organs\cite{liakos2018machine,LI2022106702,AO20221239,LI2022243,peng2024advanced}, whereas unsupervised methods—relying on self‐organizing and clustering techniques—effectively tackle segmentation tasks under annotation scarcity and complex backgrounds\cite{yan2022unsupervised,zhu2024identifying,huang2024unsupervised}. Interactive 3D segmentation methods based on the SAM are also rapidly evolving, thanks to their zero‐shot capability across diverse scenarios. SA3D\cite{cen2023segment} leverages SAM’s segmentation power by allowing the user to specify target points in a single view and then auto‐generating prompts for the next image’s mask, which are fused with NeRF information to extract the target’s 3D point cloud. 
However, when applied to 360-degree scanned rice data, such point prompting strategies face challenges due to self-occlusion characteristics. This leads to prompt point drift, resulting in model failure and inability to export point clouds. Particularly for small targets like grains, SA3D fails to achieve extraction.
SANeRF-HQ \cite{liu2023sanerf} enhances segmentation boundary accuracy in NeRF-based multi-view aggregation by leveraging density fields and RGB similarity. PointSAM \cite{zhou2024pointsam}, a widely adopted 3D annotation tool, utilizes SAM to generate extensive part-level and object-level pseudo-labels, extracting rich knowledge for fully supervised training in 3D instance segmentation. However, its single-instance-per-interaction and multi-round click optimization mechanism on point cloud data faces critical bottlenecks in interaction efficiency.

Segment Anything Model 2 (SAM2) \cite{ravi2024sam2}, a significant upgrade by Meta over the SAM, employs a unified Transformer architecture \cite{vaswani2017attention} and prompt-driven strategy. This enhanced framework not only retains zero-shot segmentation capabilities but also achieves improved segmentation accuracy and accelerated inference speeds through the integration of memory modules and optimized mask decoders. In video segmentation tasks, the model demonstrates flexible utilization of interactive prompts (e.g., points, boxes) for precise target boundary extraction, substantially enhancing operational efficiency. These advancements establish novel algorithmic design paradigms for breeding processes.


    

Although there has been considerable progress in extracting 3D organ point clouds of crops, several challenges remain:
\begin{itemize}
    \item 3D data annotation relies heavily on manual, fine‐grained labeling, which incurs high labor costs, long turnaround times, and significant subjective bias. These constraints severely hinder the creation of large‐scale, high‐precision 3D phenotyping datasets and slow the progress of crop phenotypic analysis. In the breeding field, there is an urgent need for zero‐shot, unsupervised 3D organ point‐cloud extraction solutions with robust cross‐variety adaptability.
    \item For small, self‐occluded rice grains, reconstruction-based interactive point cloud segmentation methods such as SA3D fail to extract the target grains. Therefore, it is necessary to develop a grain-level interactive point cloud extraction scheme that, with only minimal user prompts, can achieve high-quality extraction of individual grains.
    \item Mainstream interactive methods apply a single-instance, iterative correction strategy to point clouds, but their multi-click optimization mechanism creates an interaction-efficiency bottleneck. Existing approaches also exhibit significant shortcomings in triggering multi-target collaborative segmentation at the grain level with a single interaction, making them inadequate for the point cloud extraction demands of complex crops.
\end{itemize}

\begin{figure*}[ht]
    \centerline{\includegraphics[width=\linewidth]{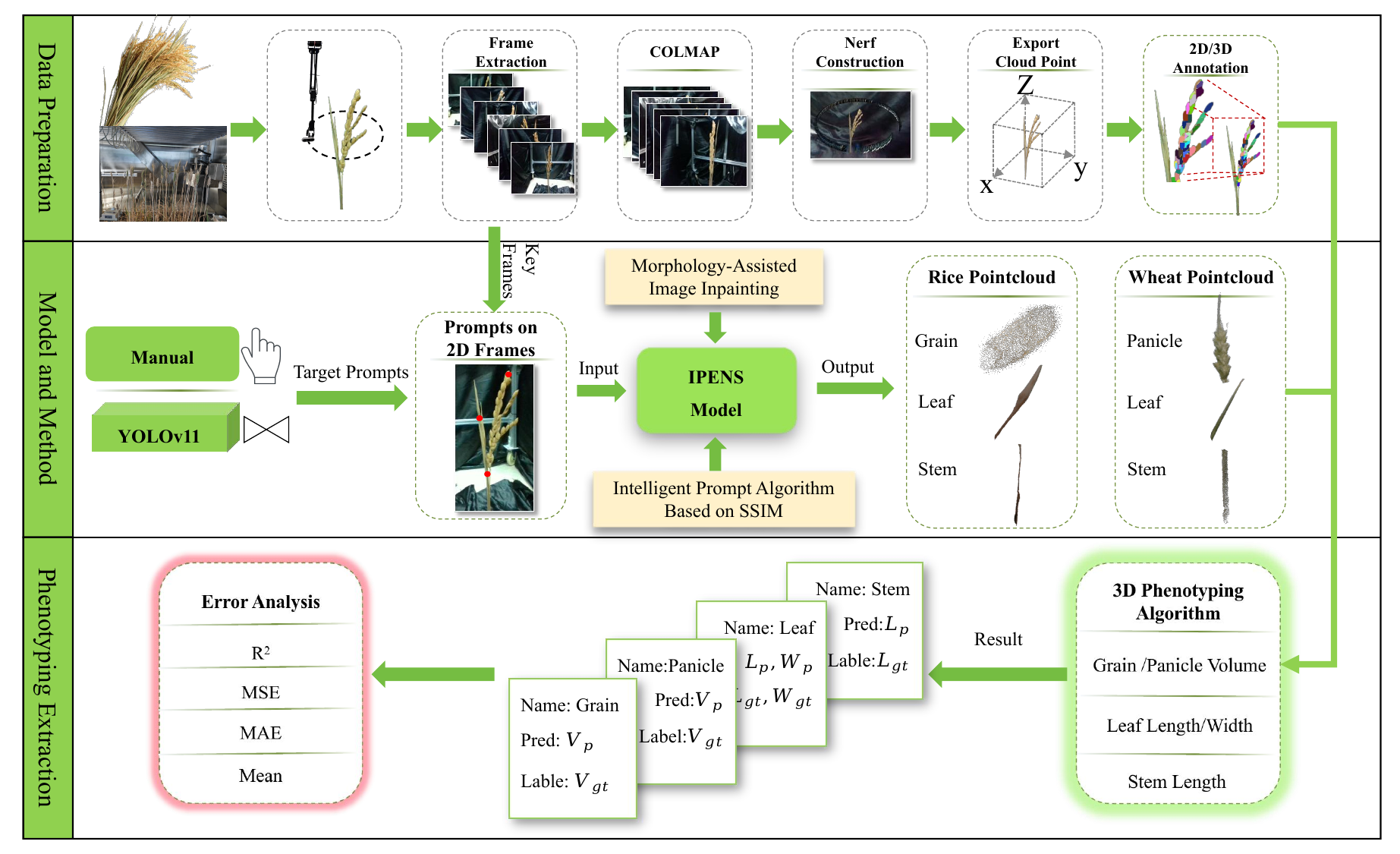}}
    \caption{Overall Workflow of IPENS: Data Preparation, Model \& Method, and Phenotyping Extraction. }
    \label{fig:workflow}
\end{figure*}

    

To address these challenges, this study introduces NeRF and SAM2 models into the field of agricultural phenotyping and proposes an interactive unsupervised prompting framework to explore their potential for high-throughput biomass phenotype extraction in rice and wheat, while also discussing the feasibility of multi-species point cloud extraction. The specific contributions are as follows:
\begin{itemize}
    \item We propose IPENS, an interactive, unsupervised, multi‐target phenotyping extraction and analysis framework based on NeRF and SAM2. 
    Breaking the traditional reliance on crop annotation, we propose a single-round prompting solution: by supplying prompt points on just a few images, leveraging SAM2’s video-propagation capability to semantically spread those prompts across the 2D image sequence, and combining NeRF’s spatial representation strengths with differentiable rendering to lift the resulting 2D masks into 3D space.
    Ultimately, high-quality 3D point-cloud extraction for rice and wheat was achieved.
    \item We propose a multi‐target collaborative extraction scheme tailored for breeding scenarios. To address the need for parallel analysis of multiple plant organs, we designed a multi‐target training strategy that allows breeders to specify prompts for several organs at once, enabling a single interaction to synchronously extract 3D point clouds of organs such as rice grains, leaves, and stems. The framework supports both manual prompting and an automated prompting scheme based on YOLOv11 \cite{khanam2024yolov11}.
    \item We designed a two-stage post-processing optimization strategy. First, morphological inpainting is applied to SAM2’s segmentation results to produce smoother, more coherent 2D masks, which in turn improves the accuracy of 3D mask extraction. Second, because self-occlusion of grains causes SAM2’s performance to degrade when targets disappear, we need to rapidly relocate their reappearance and supply new prompts to refine the segmentation. To further simplify this process for users, we introduce an intelligent prompting algorithm based on the Structural Similarity Index Measure (SSIM) \cite{wang2004image}, which helps users quickly identify and annotate rear-view frames in the video, thereby enhancing the quality of 3D point cloud extraction.
    \item Phenotypic methods were developed to compute the volume of rice grains and wheat grains, leaf surface area, and leaf length and width. The effectiveness of model in point cloud extraction was demonstrated.
    
\end{itemize}

\section{Materials and Methods}
\subsection{Overview of the Method}


Figure \ref{fig:workflow} illustrates the overall workflow. For the acquired image sequence, camera poses are first estimated using COLMAP\cite{colmap1,colmap2}, and NeRF is then employed for 3D rendering to generate the crop point cloud. 
The framework provides two interaction schemes: manual prompting or YOLOv11-assisted prompting on key frames, which are then input to the IPENS model. Two post-processing optimization strategies further enhance the 2D segmentation results and generate high-quality 3D segmentation masks.
Finally, phenotypic traits are extracted from both the manually annotated 3D point cloud and the model-generated masks, and the resulting estimation errors are evaluated.

The following sections describe, in detail, the data acquisition and reconstruction methods (Section \ref{sec:data_collection_and_reconstruction}), the dataset construction process (Section \ref{sec:data-set}), and the 3D point cloud extraction pipeline (Sections \ref{sec:ipens_model_pipeline}–\ref{sec:loss_function_implementation}). We also introduce our prompting strategy in Section \ref{sec:prompt-point} and present the two post-processing algorithms in Sections \ref{sec:residual-handling}–\ref{sec:similarity}. Finally, we outline the phenotypic methods for each organ in Section \ref{sec:rice_shape_extraction}.

\subsection{Data acquisition and reconstruction}\label{sec:data_collection_and_reconstruction}

\begin{figure}[htbp]
    \centering
    \begin{subfigure}[b]{0.23\textwidth}
        \includegraphics[width=\textwidth]{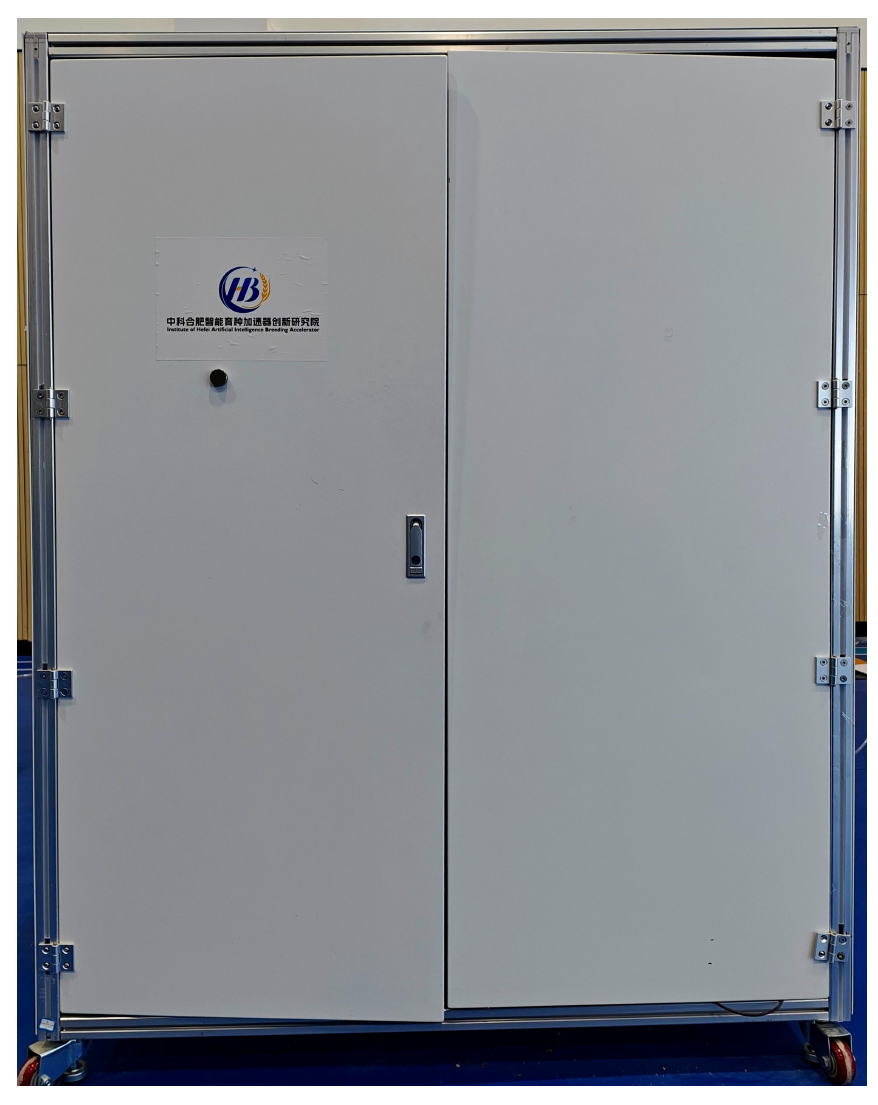}
        \caption{The exterior of the data acquisition cube}
        \label{cube-out}
    \end{subfigure}
    \hfill
    \begin{subfigure}[b]{0.23\textwidth}
        \includegraphics[width=\textwidth]{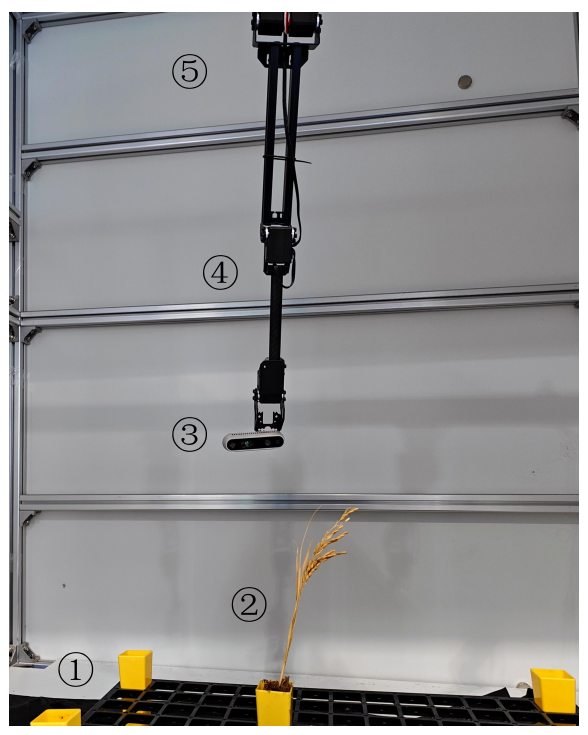}
        \caption{The interior of the data acquisition cube}
        \label{cube}
    \end{subfigure}
    \caption{Data acquisition cube. Components inside (b) include:1:A 98-cell modular seedling tray in the breeding chamber 2:Rice or other crop. 3: Intel D435i camera. 4: Robotic arm. 5: Top and side soft-box lights.}
    \label{fig:cube}
\end{figure}

Experiments were carried out on the multimodal and multitask rice (MMR) and wheat (MMW) datasets, both specifically designed for phenotyping research. To acquire high-quality 3D data, the first part of Figure \ref{fig:workflow} shows the standardized data acquisition procedure: samples were randomly selected in a breeding accelerator growth chamber, then data capture was performed inside the breeding data acquisition cube (Figure \ref{fig:cube}). A backend control system precisely guided a robotic arm to execute a 360° orbit while synchronously triggering an Intel D435i camera to record 20 s videos at 1280×720 pixels and 30 fps; key frames were then extracted at 6 fps. Next, COLMAP estimated the external camera parameters for each image, and nerfacto\cite{nerfstudio} was used for 3D reconstruction to obtain the crop’s NeRF field. Finally, the reconstructed point clouds were exported to facilitate ground-truth annotation.

\subsection{Dataset construction}\label{sec:data-set}
\begin{figure}[ht]
    \centering
    \includegraphics[width=0.5\textwidth]{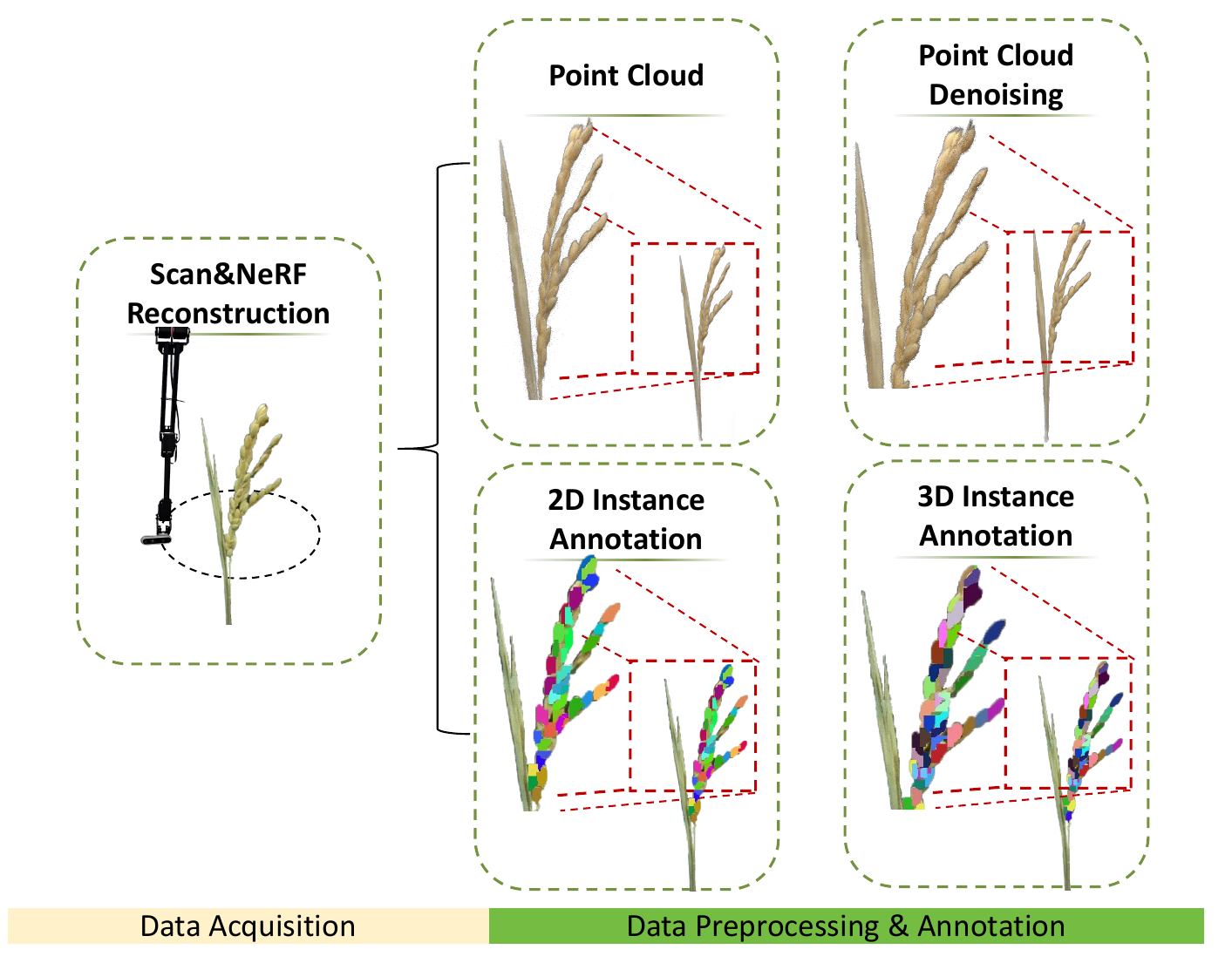}
    \caption{Data acquisition and instance segmentation }
    \label{fig:data-label}
\end{figure}

We controlled each plant’s point cloud to approximately 100,000 points and removed isolated points using a Gaussian‐distribution filter in Open3D. Next, precise instance labeling was performed with CloudCompare. To further enhance data quality, we applied the Gaussian‐distribution denoising method again to each individually labeled instance, effectively eliminating errors introduced by labeling inaccuracies.

As shown in Figure \ref{fig:data-label}, the MMR and MMW datasets comprise multiview crop images, 2D instance segmentation masks, raw point‐cloud data (XYZRGB), and annotated semantic- and instance-segmented point clouds. The 3D instance statistics are presented in Table \ref{tab:data_statistics}.
\begin{table}[H]
    \centering
    \caption{3D instance statistics results}
    \label{tab:data_statistics}
    \begin{tabular}{c c c c}
        \toprule
\textbf{Dataset} & \textbf{Plants} & \textbf{Class}  & \textbf{Instance Seg} \\
\midrule 
\multirow{3}{*}{MMR} &\multirow{3}{*}{86}& Grain & 3316  \\
&     & Leaf  & 94  \\
&     & Stem  & 128   \\
\hline
\multirow{3}{*}{MMW} &\multirow{3}{*}{35}& Panicle & 35  \\
&     & Leaf  & 119  \\
&     & Stem  & 36   \\
\bottomrule
\end{tabular}
\end{table}

\subsection{IPENS model pipeline}\label{sec:ipens_model_pipeline}
\begin{figure*}[htbp]
    \centerline{\includegraphics[width=0.95\linewidth]{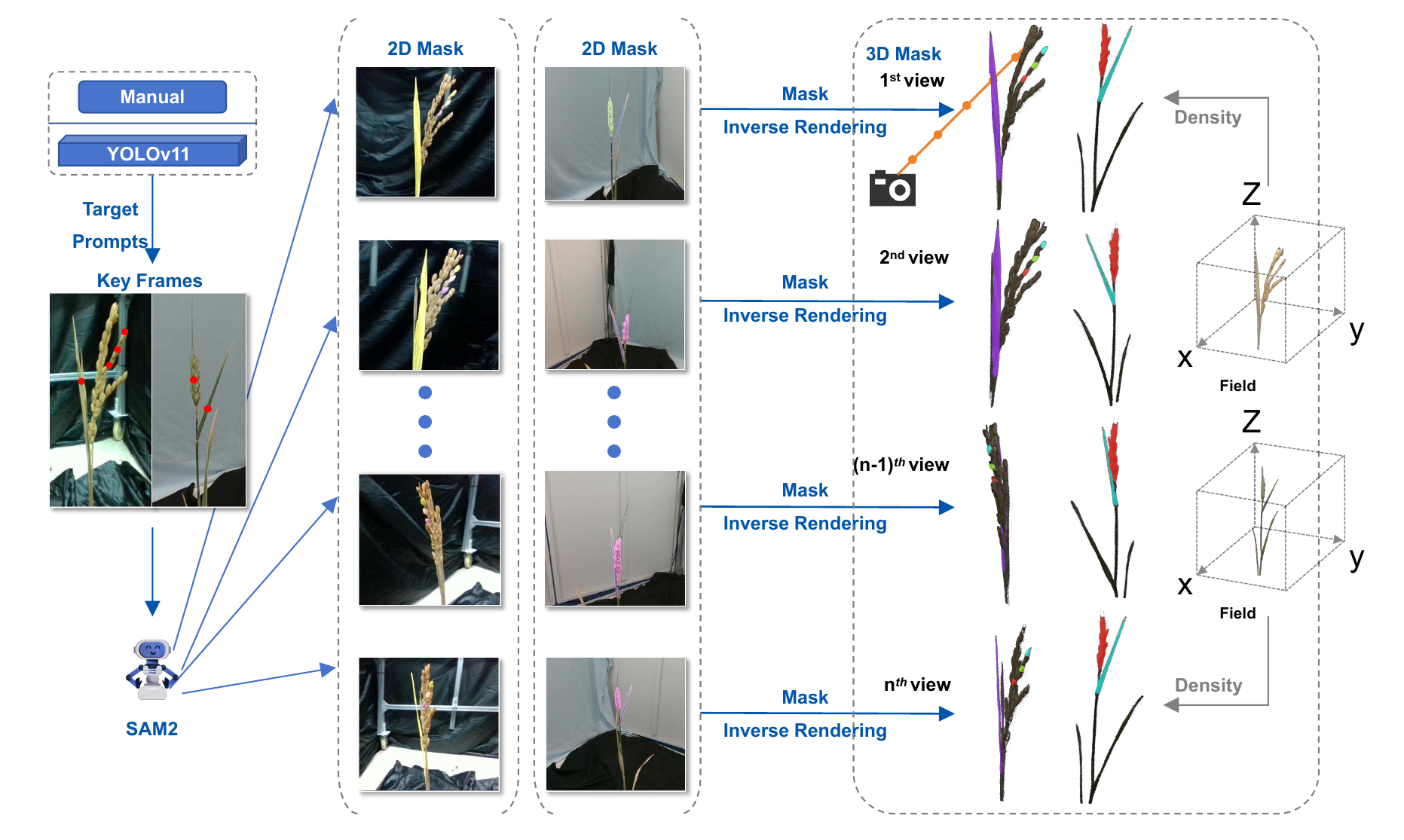}}
    \caption{IPENS Model. Given a radiance field trained on rice or wheat, the model first takes manual inputs or YOLO prompts as input. It then uses SAM2 to generate 2D masks for the image sequence. Based on the radiance field information, a mask inverse rendering process is performed iteratively, ultimately obtaining the 3D masks.}
    \label{fig:model}
\end{figure*}

To address interactive phenotyping requirements, as shown in the second part of Figure \ref{fig:workflow}, IPENS provides two types of prompts: (1) Interactive prompts, where users can select key frames from the video (with at least the first frame) and manually provide multiple targets, which can be prompt points or bounding box coordinates for objects such as grains, panicles, leaves, and stems; (2) To further reduce user interaction costs, this paper also proposes an automatic prompt method based on YOLOv11 to assist in generating target prompt information. Specifically, we use a YOLOv11 model pretrained on only five crops for instance segmentation of 2D images to assist in generating the prompt information. Compared to previous versions, the YOLOv11 model not only achieves higher segmentation accuracy but also has faster inference speed, capable of accurately locating the centers and boundaries of targets. Since SAM2 does not require highly precise prompt information, providing a general segmentation area is sufficient for prompts, so YOLOv11 is trained with very little data.

Figure \ref{fig:model} illustrates the core point‐cloud extraction process of IPENS. Prompt information is fed into SAM2 for 2D image segmentation. Thanks to SAM2’s streaming‐memory mechanism, segmentation masks propagate continuously and spatiotemporally consistently, significantly reducing user interaction and improving real‐time performance. With minimal intervention, high‐quality 2D segmentation masks of the target across multiple views are obtained.

Since the crop’s NeRF field has already been constructed after data acquisition, we leverage it to introduce a multi‐target collaborative optimization strategy. By applying mask inverse rendering, we project the 2D masks of multiple targets into 3D space frame by frame, forming an initial coarse 3D mask. This process is iteratively executed across all views. As the number of iterations increases, the 3D mask’s accuracy progressively improves, ultimately yielding a high‐quality 3D mask for the targets.

Following this process, users can interact with IPENS in a single round, clicking on several different targets within the key frames to obtain the corresponding 3D masks.

\subsection{3D Mask Representation}\label{sec:3d-mask-representation}
For a 3D voxel grid $V \in \mathbb{R}^{L \times W \times H}$ containing multiple objects, its rendering result onto the two-dimensional plane can be represented as
\begin{equation}
    M(r) \;=\; \int_{t_n}^{t_f} w\bigl(r(t)\bigr)\,G\bigl(r(t)\bigr)\,\mathrm{d}t, 
\end{equation}
where $r(t)$ denotes the position of the ray at the parameter $t$, $w\bigl(r(t)\bigr)$ is the weight function of the ray at that position, and $G\bigl(r(t)\bigr)$ is defined as
\begin{equation}
    G\bigl(r(t)\bigr) \;=\; 
    \begin{bmatrix}
    V_{o_1}\bigl(r(t)\bigr) \\
    V_{o_2}\bigl(r(t)\bigr) \\
    \vdots \\
    V_{o_n}\bigl(r(t)\bigr)
    \end{bmatrix},
\end{equation}
where $V_{o_i}\bigl(r(t)\bigr)$ represents the mask score of the $i_{th}$ object obtained from the voxel grid $V$ at the position $r(t)$.

\subsection{Loss Function Implementation}\label{sec:loss_function_implementation}
The purpose of mask inverse rendering is to project the 2D masks of the image $I$ into the 3D space. Let $M^{\mathrm{SAM2}}$ denote the mask generated by SAM2, and $M_i$ represent the mask of the $i_{th}$ target.

\begin{equation}
    \begin{split}
    L_{\mathrm{proj}} 
    &= -\sum_{n} \sum_{r \in R(I)} 
    M_{i}^{\mathrm{SAM2}}(r)\cdot M_{i}(r) \\
    &\quad + \lambda \sum_{n} \sum_{r \in R(I)}
    \Bigl[ \bigl(1 - M_{i}^{\mathrm{SAM2}}(r)\bigr) \cdot M_{i}(r) \Bigr].
    \end{split}
\end{equation}
where $R(I)$ denotes the ray set of the image $I$,
$M_{i}^{\mathrm{SAM2}}(r)$ is the value of the $i_{th}$ mask generated by the SAM2 model at position $r$,
$M_i(r)$ is the mask value of the true target at position $r$, and $\lambda$ is a balancing weight hyperparameter.
The model continuously updates the mask score of $M(r)$. Finally, it fuses the ID, mask score, and point cloud color to export targets in different colors.

\subsection{Prompting strategy}\label{sec:prompt-point}
Before SAM2 segmentation, users often click on the center of the target to obtain a positive sample prompt point and select around the target as negative sample prompt points. However, this strategy of prompt points can lead to an imbalance between positive and negative samples, resulting in excessive encroachment of the background area on the target. This can be manually controlled during manual prompting. In the case of YOLO assistance, to automatically solve the problem of positive and negative sample imbalance and further improve the accuracy of target segmentation, this paper proposes a "Positive-Negative-Positive" strategy, which makes full use of negative samples while generating more positive prompt points to balance the ratio of positive and negative samples. Specifically, for each target instance, YOLO is first used to determine its center point $(c_X, c_Y)$. To suppress the background area and improve the precision of the target area, the algorithm selects five adjacent center points around the target center as negative sample prompt points. Through these negative sample prompt points, the model can distinguish between the target and the background, reducing the false detection rate. Then, uniformly distributed sampling points are constructed near the center point as positive prompt points. This method adopts fixed grid sampling, where the preset grid size is 3 and the sampling range (radius) is 1. Taking the y-axis direction as an example, for the upper and lower neighborhoods of the center point, by looping through the offsets from $-radius$ to $radius$ (with a step size of the grid size), the OpenCV function \texttt{cv2.pointPolygonTest} is used to judge whether the generated sampling points are inside the target polygon. Only points that meet the conditions are retained as positive prompt points.

This method not only balances the ratio of positive and negative samples but also effectively improves the segmentation accuracy of the SAM2 model, thereby more accurately recovering the complete shape of the target.

\subsection{Residual handling}\label{sec:residual-handling}
Due to factors such as illumination, occlusion, and background noise, 2D organ segmentation results often suffer from incompleteness and discontinuities, affecting the accurate extraction of subsequent 3D data. To address this issue, this paper adopts a post-processing method based on morphological operations, optimizing the segmentation mask by combining closing and opening operations. Specifically, algorithm \ref{alg:postprocess} first processes the binarized segmentation mask using morphological closing. This process, which involves dilation followed by erosion, effectively fills holes within the target area and repairs discontinuities caused by incompleteness; subsequently, morphological opening is applied to the result of the closing operation to further remove noise. This operation, which involves erosion followed by dilation, can eliminate fine noise points that may be introduced by the closing operation, thereby obtaining a smoother and more coherent segmentation result. This post-processing scheme improves the 2D segmentation accuracy of crop organs, thereby enhancing the accuracy of subsequent 3D point cloud extraction.
\begin{algorithm}[ht]
\caption{Residual handling}\label{alg:postprocess}
\begin{algorithmic}
\STATE \textbf{Input:}Segmentation mask $M \in \mathbb{R}^{ H \times W}$
\STATE \textbf{Output:} Post-processed binary mask $\tilde{M} \in \mathbb{R}^{ H \times W}$
\FOR{each pixel $p$ in $M$}
    \IF{$p > 0$}
        \STATE $p \gets 255$ \COMMENT{Set as foreground}
    \ELSE
        \STATE $p \gets 0$ \COMMENT{Set as background}
    \ENDIF
\ENDFOR
\STATE $K \gets \text{ones}(5,5)$ \COMMENT{Define a $5\times5$ structuring element}
\STATE $M \gets \operatorname{MorphologicalClose}(M, K)$ \COMMENT{Closing operation: fill holes within the target}
\STATE $\tilde{M} \gets \operatorname{MorphologicalOpen}(M, K)$ \COMMENT{Opening operation: remove noise}
\STATE $\tilde{M} \gets \tilde{M} / 255$ \COMMENT{Normalize to $[0,1]$}
\STATE $\tilde{M} \gets \operatorname{Reshape}(\tilde{M}, (1,H,W))$ \COMMENT{Adjust output dimensions}
\end{algorithmic}
\end{algorithm}

\subsection{A fast algorithm for finding prompt frames based on SSIM}\label{sec:similarity}
\begin{figure}[htbp]
    \centerline{\includegraphics[width=\linewidth]{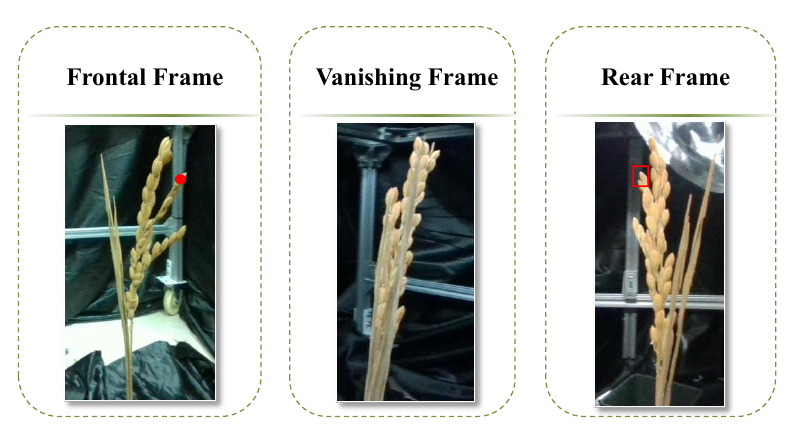}}
    \caption{Target Vanishing to Reappearance Process}
    \label{fig:SSIM}
\end{figure}
Due to the self-occlusion problem, SAM2 often experiences a decrease in segmentation accuracy for targets in subsequent images when the target is occluded, i.e., when the target disappears from view, as shown in Figure \ref{fig:SSIM}. To automatically correct point cloud results and further improve the segmentation accuracy of the model, it is necessary to provide prompts for rear frames.
Therefore, to identify rear frames and improve detection efficiency under a large amount of image data, this paper proposes a fast decision algorithm based on SSIM and parallel computing.
Specifically, Algorithm \ref{alg:Rear-view } first preprocesses the front reference frame, including converting it to a grayscale image and scaling it to a lower resolution, to reduce the complexity of subsequent SSIM calculations. Then, parallel processing is applied to the set of images to be detected. For each image, it is first preprocessed in the same way as the reference frame, and then its horizontal mirror image is generated. Subsequently, the SSIM value $s_{\text{raw}}$ between the preprocessed original image and the reference frame and the SSIM value $s_{\text{mirror}}$ between the horizontal mirror image and the reference frame are calculated separately. If the difference $s_{\text{mirror}}-s_{\text{raw}}$ exceeds a preset threshold, the image is considered a rear frame. Finally, by counting the indices of all images detected as rear frames, the minimum and maximum indices are taken as the positions of the first and last occurrences of rear frames, respectively. This method significantly reduces computation time by utilizing downsampling and parallel processing techniques while maintaining high detection accuracy.
\begin{algorithm}[ht]
    \caption{Find Rear-view Frames}\label{alg:Rear-view }
    \begin{algorithmic}
    \STATE \textbf{Input:}\texttt{reference frame and images} 
    \STATE \textbf{Output:}\texttt{first and last indices}
    \STATE $pre\_ref \gets \operatorname{Preprocess}(reference)$
    \STATE $files \gets \operatorname{SortedList}(images)$
    \STATE $indices \gets \varnothing$
    \FOR{$(idx, f) \in files$ \textbf{in parallel}}
        \STATE $img \gets \operatorname{ReadImage}(f)$
        \STATE $proc \gets \operatorname{Preprocess}(img)$
        \STATE $mirror \gets \operatorname{Preprocess}(\operatorname{HorizontalFlip}(img))$
        \STATE $s\_raw \gets \operatorname{SSIM}(pre\_ref, proc)$
        \STATE $s\_mirror \gets \operatorname{SSIM}(pre\_ref, mirror)$
        \IF{$(s\_mirror - s\_raw) > threshold$}
            \STATE Add $idx$ to $indices$
        \ENDIF
    \ENDFOR
    \IF{$indices \neq \varnothing$}
        \STATE $first \gets \min(indices)$
        \STATE $last \gets \max(indices)$
    \ENDIF
    \end{algorithmic}
    \end{algorithm}

\subsection{Phenotypic Data Extraction Method}\label{sec:rice_shape_extraction}
To accurately compute the phenotypic results of target organs, it is necessary to preprocess the point cloud extracted by the model. This specifically includes setting the voxel size to 0.01, performing voxel downsampling, and noise reduction; using the Alpha Shapes algorithm for initial mesh construction, followed by hole filling and loop subdivision on the mesh to generate multiple triangular facets.
\subsubsection{Extraction of volumes for grains and panicles}
Calculate the point cloud volume $V$ using the following formula based on the voxel size and the number of points in the point cloud $num\_points$, to obtain the volumes of grains and panicles, facilitating subsequent indicator calculations.
\begin{equation}
    V = num\_points * 0.01^3
\end{equation}

\subsubsection{Leaf surface area}
The mesh consists of a series of triangular facets, and the leaf area is approximated by calculating the sum of the areas of all triangles. 
The three vertices of each triangle are represented as vectors
$ \mathbf{v_1},\; \mathbf{v_2},\; \mathbf{v_3}.$
The two edge vectors are
\begin{equation}
    \mathbf{a} = \mathbf{v_1} - \mathbf{v_2}, 
    \quad
    \mathbf{b} = \mathbf{v_1} - \mathbf{v_3}.
\end{equation}
Calculate the cross product of vectors $\mathbf{a}$ and $\mathbf{b}$ to obtain a vector $\mathbf{c} = \mathbf{a} \times \mathbf{b}$ that is perpendicular to the plane of the triangle.
Calculate half the magnitude of the cross product as the area of the triangle:
\begin{equation}
    A_{\triangle} \;=\; 0.5 \;\times\; \|\mathbf{c}\|.
\end{equation}
Sum the areas of all triangles to obtain the surface area of the mesh:
\begin{equation}
    A_{\text{mesh}} \;=\; \sum A_{\triangle}.
\end{equation}

\subsubsection{Leaf Length and Width}
To calculate the length, a point set is extracted based on the nearest neighbor algorithm to fit the midrib of the leaf. The length of the curve formed by the point set represents the leaf length. Specifically,
$s_{1}$: Let the preprocessed leaf point cloud be $\mathcal{P} = \{\mathbf{p}_i \in \mathbb{R}^3\}_{i=1}^N$, where $\mathbf{p}_i = (x_i, y_i, z_i)^\top$. Perform principal component analysis (PCA) on it and calculate the covariance matrix:

\begin{equation}
    \mathbf{C} = \frac{1}{N}\sum_{i=1}^N (\mathbf{p}_i - \bar{\mathbf{p}})(\mathbf{p}_i - \bar{\mathbf{p}})^\top ,
\end{equation}
where $\bar{\mathbf{p}}$ is the centroid of the point cloud. Through eigenvalue decomposition $\mathbf{C} = \mathbf{V}\mathbf{\Lambda}\mathbf{V}^\top$, the main direction $\mathbf{v}_1$ (corresponding to the largest eigenvalue $\lambda_1$) is obtained.
Treat the two farthest endpoints found on the first principal component axis as the starting and ending points for calculating the leaf length.
\begin{equation}
    \begin{cases}
        \mathbf{e}_s = \arg\min_{\mathbf{p}\in\mathcal{P}} \mathbf{p}^\top\mathbf{v}_1 \\
        \mathbf{e}_t = \arg\max_{\mathbf{p}\in\mathcal{P}} \mathbf{p}^\top\mathbf{v}_1
        \end{cases} 
\end{equation}

$s_{2}$: Build a KD-tree  $\mathcal{T} = \text{BuildKDTree}(\mathcal{P})$.
Starting from the starting point, perform K-nearest neighbor (KNN) search. Sort the K candidate points in a greedy manner based on their distance from the starting point, and execute the loop process until the preset conditions are met.

$s_{3}$: Let the current base point be $\mathbf{b}_k$, and the candidate point set be $\mathcal{N}_k = \{\mathbf{q}_j\}_{j=1}^K$, satisfying $\mathcal{N}_k = \text{KNN}(\mathcal{T}, \mathbf{b}_k, K)$. Calculate the angle between the vector formed by the next base point and $\mathbf{b}_k$ and the principal component axis
\begin{equation}
    \theta_j = \arccos\left(\frac{(\mathbf{q}_j - \mathbf{b}_k)^\top \mathbf{v}_1}{\|\mathbf{q}_j - \mathbf{b}_k\|_2 \cdot \|\mathbf{v}_1\|_2}\right) < \theta_{\max},
\end{equation}
where $\theta_{\max} = \pi/2$ is the maximum allowable deviation angle. 
If $\theta_j$ exceeds the maximum angle $\pi/2$, skip it; otherwise, select it. If no points meet the criteria, return the point with the smallest angle.

$s_{4}$: Termination conditions:
\begin{equation}
     \|\mathbf{b}_k - \mathbf{e}_t\|_2 > \epsilon \ \text{and} \ k < M_{\max},
\end{equation}
where $\epsilon$ is the positional tolerance and $M_{\max}$ is the maximum number of base points. 
If enough base points are found or the current base point coincides with the endpoint, the loop ends.

$s_{5}$: Downsample the collected base point set $\mathcal{B} = \{\mathbf{b}_m\}_{m=1}^M$ to obtain a set of base points along the curved path of the leaf. Summing the distances between consecutive points gives the total length of the leaf.
\begin{equation}
L = \sum_{m=1}^{M-1} \|\mathbf{b}_{m+1} - \mathbf{b}_m\|_2.
\end{equation}

The leaf width is calculated based on a mesh parameterization method that flattens the leaf in the width direction, followed by 2D interpolation. The interval width is defined as the distance between boundary points of each row, and the length of the widest interval represents the leaf width. Specifically, define the triangular mesh: $\mathcal{M} = (\mathcal{V}, \mathcal{E}, \mathcal{F})$, where $\mathcal{V} = \{\mathbf{v}_i \in \mathbb{R}^3\}$ is the set of vertices, $\mathcal{E}$ is the set of edges, and $\mathcal{F}$ is the set of facets.

$s_{1}$: Mesh parameterization allows the transformation of the leaf's 3D mesh into a 2D space by minimizing energy loss. This paper employs the As-Rigid-As-Possible (ARAP) algorithm to create mapping connections between mesh vertices in both 3D and parameter spaces. The re-meshed 2D leaf grid maintains a one-to-one mapping relationship and shares the topological structure of the 3D mesh model vertices, enabling mutual retrieval based on the mapping relationship. 
To solve the mapping $\phi: \mathcal{V} \rightarrow \mathbb{R}^2$, minimize the energy: 
\begin{equation} 
E_{\text{ARAP}} = \sum_{(i,j)\in\mathcal{E}} w_{ij} \|\phi(\mathbf{v}_i) - \phi(\mathbf{v}_j) - \mathbf{R}_i(\mathbf{v}_i - \mathbf{v}_j)\|_2^2,
\end{equation}
where $\mathbf{R}_i \in SO(2)$ is the local rotation matrix, and $w_{ij}$ is the cotangent weight.

$s_{2}$: Let the parameterized coordinates be $\{\mathbf{u}_i = (u_i, v_i)^\top\}$. 
Calculate the principal component direction of the parameterized coordinates:
\begin{equation}
\mathbf{w} = \arg\max_{\|\mathbf{a}\|=1} \text{Var}(\{\mathbf{u}_i^\top \mathbf{a}\}).
\end{equation}

$s_{3}$: Perform 2D point interpolation.
Define sampling intervals on the main axis direction $\mathbf{w}$:
\begin{equation}
   s_k = s_{\min} + \frac{k}{n}(s_{\max} - s_{\min}), \quad k=0,1,...,n
\end{equation}
where $s_{\min} = \min_i \mathbf{u}_i^\top \mathbf{w}$, $s_{\max} = \max_i \mathbf{u}_i^\top \mathbf{w}$.
For each sampling position $s_k$, search for boundaries along the normal direction $\mathbf{w}^\perp = [-w_y, w_x]^\top$:
\begin{equation}
    \begin{cases}
        t_k^l = \min_{\mathbf{u}_i^\top \mathbf{w}=s_k} \mathbf{u}_i^\top \mathbf{w}^\perp \\
        t_k^r = \max_{\mathbf{u}_i^\top \mathbf{w}=s_k} \mathbf{u}_i^\top \mathbf{w}^\perp 
        \end{cases}
\end{equation}

$s_{4}$: Calculate the leaf width: For each row of interpolated boundary points, compute the Euclidean distance in the width direction $W_k = t_k^r - t_k^l$. 
Take the maximum value across all rows as the leaf width
\begin{equation}
W = \max_{1 \leq k \leq n} W_k .
\end{equation}

\section{Experiments}
This section quantitatively evaluates the segmentation performance of IPENS on MMR and MMW, and analyzes the phenotypic results. It further demonstrates that IPENS can perform high-quality instance segmentation of multiple rice or wheat organs in an unsupervised interactive manner.

\subsection{Experimental Details}
To avoid the impact of different experimental conditions on the model, all experiments in this study were conducted under the same environment. The software and hardware configurations of the platform are shown in Table \ref{tab:environment}.
\begin{table}[ht]
    \centering
    \small
    \begin{tabular}{lll}
        \toprule
        \textbf{Environment} & \textbf{Type} & \textbf{Name} \\
        \midrule
        \multirow{4}{*}{\textbf{Hardware}} 
            & CPU  & i9-12900K \\
            & GPU  & RTX 4090D \\
            & RAM  & 64 GB \\
            & VRAM & 24 GB \\
        \midrule
        \multirow{3}{*}{\textbf{Software}} 
            & OS    & Ubuntu 22.04 \\
            & CUDA & 12.4 \\
            & Python & 3.9 \\
        \bottomrule
    \end{tabular}
    \caption{System Configuration}
    \label{tab:environment}
\end{table}
Nerfacto iterates 30,000 steps to build the radiance field. During the segmentation process, the SGD optimizer is used with a learning rate of 0.1.
The mask for a single view is rendered in chunks. Due to the limitations of video memory size, the trade-off between time and space needs to be considered. We tested the execution time for each chunk with the number of rays ranging from $2^{11}$ to $2^{19}$ per single viewpoint, as shown in Table \ref{tab:batch_ray_perf}. During training, using a larger number of rays may make memory management more efficient, better utilizing the parallel computing capabilities of the GPU, thereby reducing the memory allocation overhead per ray. Therefore, we choose to set $2^{14}$ rays in each chunk.
\begin{table}[ht]
\centering
\small  
\setlength{\tabcolsep}{3pt}  
\caption{Time and video memory conditions under different numbers of rays per batch}
\label{tab:batch_ray_perf}
\begin{tabular}{lrrrr}
\toprule
\textbf{Config} & \textbf{Batch Rays} & \textbf{Time (s)} & \textbf{VRAM (MiB)} \\
\midrule
1$\ll$11 & 2048   & 158.6   & 14836 \\
1$\ll$12 & 4096   & 96.2    & 10896 \\
1$\ll$13                  & 8192   & 70.1    & 9158  \\
1$\ll$14                  & 16384  & 69.3  & 8736  \\
1$\ll$15                  & 32768  & 79.4  & 9232  \\
1$\ll$16                  & 65536  & 79.8  & 11140 \\
1$\ll$17                  & 131072 & 82.3  & 12871 \\
1$\ll$18                  & 262144 & 83.3  & 23368 \\
1$\ll$19                  & 524288 & --   & OOM   \\
\bottomrule
\end{tabular}
\end{table}

To maximize the accuracy of point cloud extraction, this experiment employs the recently released $sam2.1\_hiera\_large$ model. Given that the core objective of this paper is to reduce users' time costs, and considering that data annotation and fine-tuning of large models are time-consuming tasks, 
no fine-tuning was performed on the data in the experiment.



Common 3D segmentation evaluation metrics were used: IoU, Precision, Recall, F1-score, and average inference time.
For each instance, IoU is calculated by computing the intersection of the predicted region with the true region for that instance, then dividing it by their union. It measures the overlap between the predicted instance region and the true instance region, ranging from [0, 1], with higher values indicating closer alignment between prediction and truth. mIoU represents the ability to correctly segment 3D point clouds into various categories.
The formulas are as follows:
\begin{equation}
    \text{IoU}_i = \frac{|P_i \cap T_i|}{|P_i \cup T_i|}
\end{equation}
where: $P_i$ is the region predicted by the model for the $i_{th}$ instance, and $T_i$ is its true region.
\begin{equation}
    \text{mIoU} = \frac{1}{N} \sum_{i=1}^{N} \text{IoU}_i
\end{equation}
where $N$ is the total number of categories.

Precision represents the proportion of true positive samples among those predicted as positive, indicating the reliability of the prediction results.
\begin{equation}
    \text{Precision}_i = \frac{|P_i \cap T_i|}{|P_i|}
\end{equation}

Recall represents the proportion of the true region that is covered by the prediction. A higher value indicates fewer missed detections.
\begin{equation}
    \text{Recall}_i = \frac{|P_i \cap T_i|}{|T_i|}
\end{equation}

F1-score is the harmonic mean of precision and recall, providing a comprehensive measure of model performance. A higher value indicates better overall performance of the model.
\begin{equation}
    \text{F1}_i = 2 \times \frac{\text{Precision}_i \times \text{Recall}_i}{\text{Precision}_i + \text{Recall}_i}
\end{equation}

Average inference time is a common speed metric used to evaluate model performance, representing the average time consumed by the model for one instance segmentation prediction.
Assume that the model makes predictions for \(M\) samples. The formula for calculating the average inference time is:
\begin{equation}
    T_{\text{avg}} = \frac{1}{M} \sum_{j=1}^{M} T_j
\end{equation}
where \(T_j\) is the inference time for the $j_{th}$ prediction.

To evaluate the phenotypic quality of the segmented targets, we compared the predicted results with the ground truth. Root Mean Square Error (RMSE), Mean Absolute Error (MAE), and R-squared (R²) were used. These metrics reflect the deviation between the model's predictions and the true values from different perspectives, as well as the goodness of fit of the model. They provide important references for model optimization and comparison.

RMSE measures the overall prediction error of the model by calculating the square root of the average of the squared residuals between the predicted and true values. RMSE is sensitive to outliers, as larger errors are significantly magnified due to the squaring operation. The formula is:
\[
\text{RMSE} = \sqrt{\frac{1}{n}\sum_{i=1}^{n}(y_i - \hat{y}_i)^2}
\]
where \( y_i \) is the true value, \( \hat{y}_i \) is the predicted value, and \( n \) is the total number of samples.

MAE directly calculates the average of the absolute differences between the predicted and true values. It is suitable for scenarios with noisy data or uniformly distributed errors. The formula is:
\[
\text{MAE} = \frac{1}{n}\sum_{i=1}^{n}|y_i - \hat{y}_i|
\]

R² evaluates the goodness of fit of the model by explaining the proportion of variance in the dependent variable. Its range is [0,1], and a value closer to 1 indicates better model fit. The definition is:
\[
\text{R}^2 = 1 - \frac{\sum_{i=1}^{n}(y_i - \hat{y}_i)^2}{\sum_{i=1}^{n}(y_i - \bar{y})^2}
\]

\subsection{Quantitative Experiment}

\subsubsection{The Effectiveness of IPENS}



In this study, to verify the effectiveness of IPENS, 30\% of the data set was taken as the validation set, and the remaining part was provided as the training set for the comparison algorithm training.

To demonstrate the segmentation capability of IPENS, this experiment will take the crop organs observable in the first frame for testing.
Prompting is performed on the first frame and the rear frame of the video, manually prompting 2 positive samples and 2 negative sample prompt points, which are submitted to the model for 3D segmentation. The IoU, precision, recall, and F1 score indicators are quantitatively evaluated according to the ground truth in the data set.

\begin{figure}[h]
    \centerline{\includegraphics[width=\linewidth]{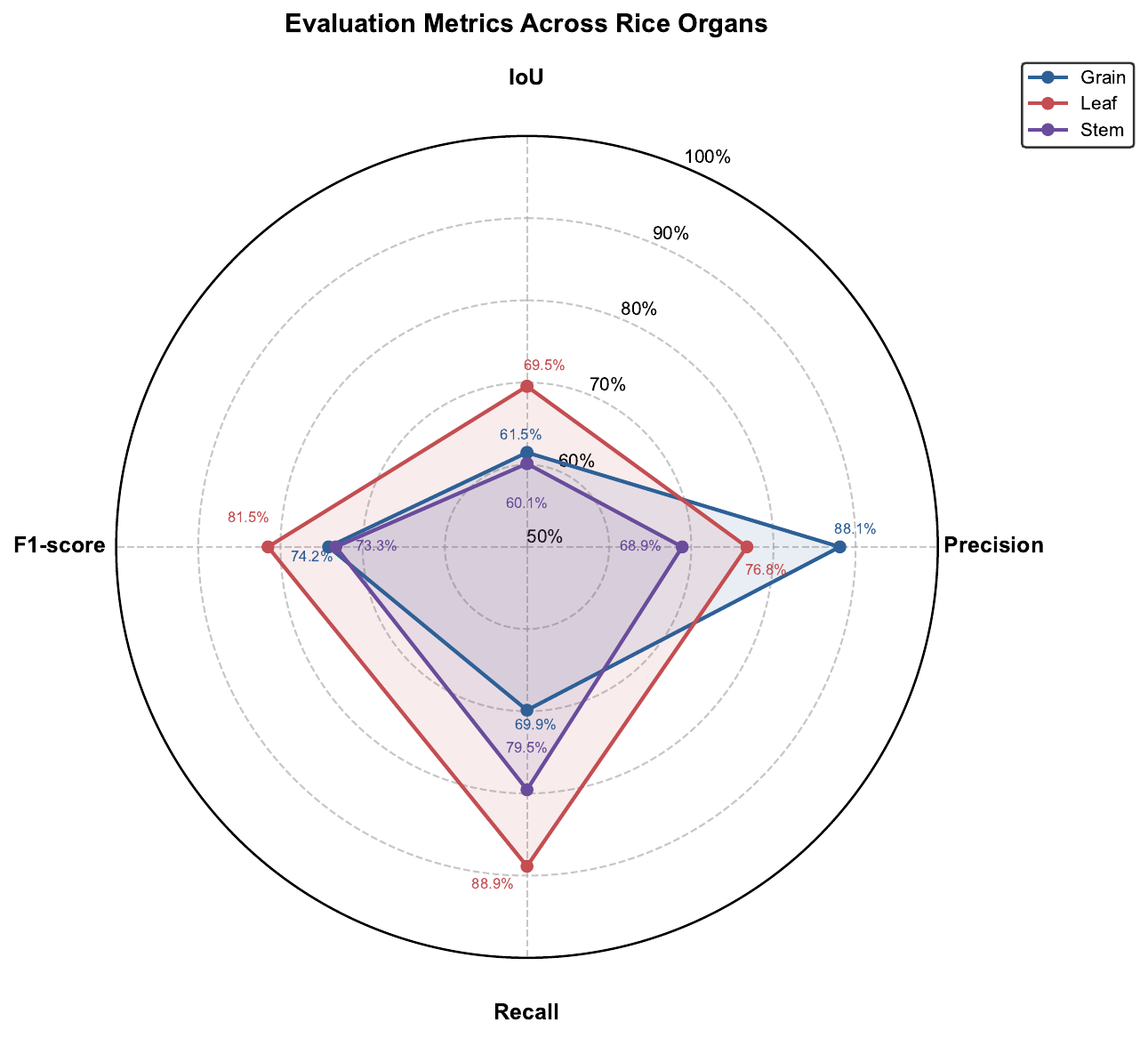}}
    \caption{Radar Chart of Evaluation Metrics Across Rice Organs}
    \label{fig:radar}
\end{figure}

The results of the model on the MMR are shown in Table \ref{tab:metrics_percent}, and Figure \ref{fig:radar} is the visualization of the metrics. In terms of the IoU index, the segmentation accuracy of Grain, Leaf, and Stem is 61.48\%, 69.54\%, and 60.13\%, respectively. The high Precision and low Recall of grain are due to the regular shape and obvious color features of grain, which enable the model to accurately identify positive samples, but some occlusions or illumination changes lead to missed detections. The leaf segmentation effect is the best, which is related to the regular shape of the leaves and the high color contrast. The low precision indicates that there are cases where stems are mistakenly judged as leaves. The overall performance of the stem is weak, which may be due to the slender morphological characteristics of the stem increasing the difficulty of boundary localization or the blurred boundary with the leaf leading to feature confusion. It is necessary to optimize the model's ability to recognize the stem.

\begin{table}[htbp]
    \centering
    \caption{Evaluation Metrics for Different Rice Organs (Percentage)}
    \label{tab:metrics_percent}
    \setlength{\tabcolsep}{3pt} 
    \begin{tabular}{lcccc}
    \toprule
    Organ   & IoU    & Precision  & Recall   & F1-score \\
    \midrule
    Grain   & 61.48      & 88.08         & 69.87       & 74.15        \\
    Leaf    & 69.54      & 76.76         & 88.86       & 81.52        \\
    Stem    & 60.13      & 68.89         & 79.55       & 73.27        \\
    \bottomrule
\end{tabular}
\end{table}

\begin{figure}[h]
    \centerline{\includegraphics[width=\linewidth]{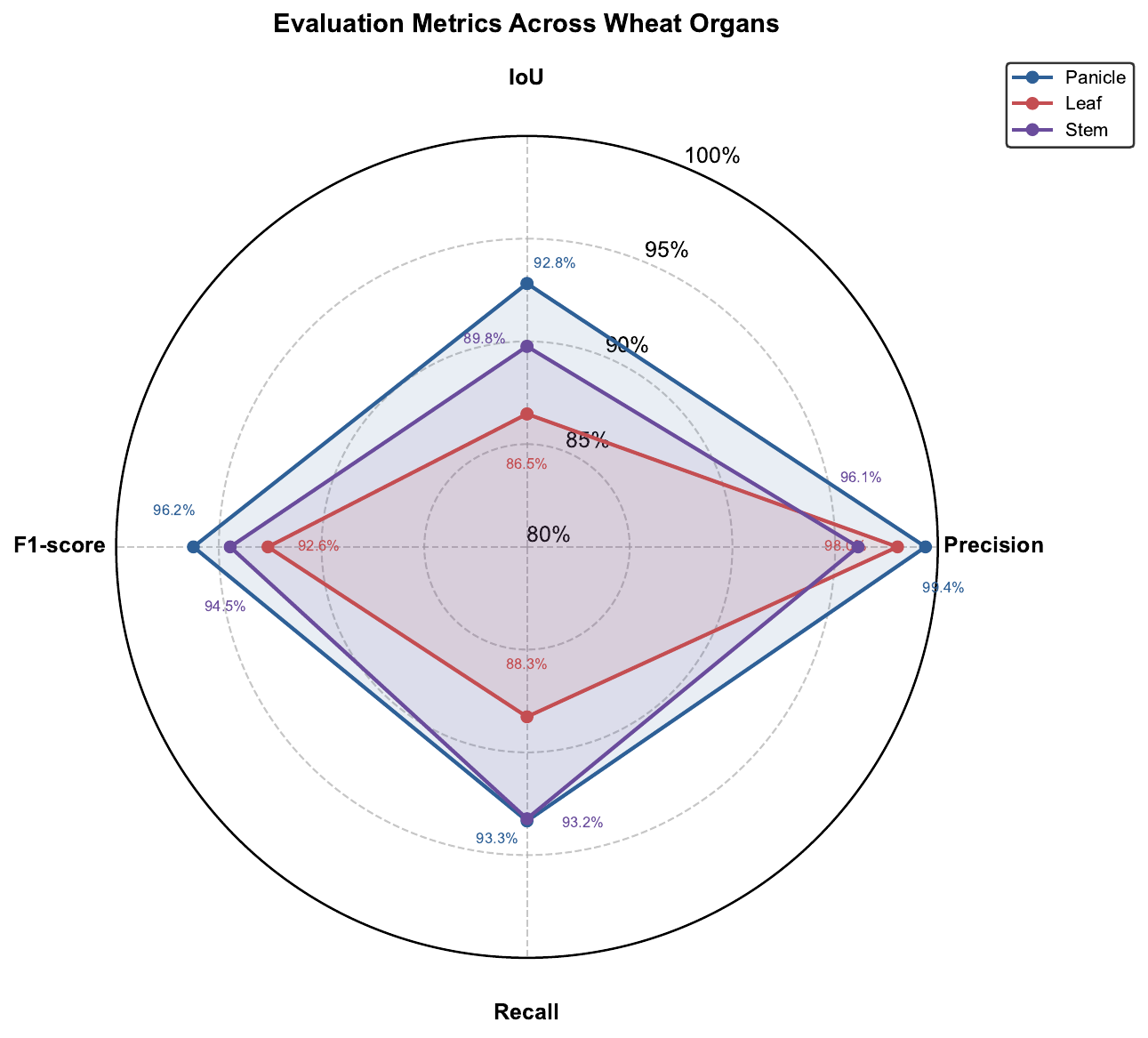}}
    \caption{Radar Chart of Evaluation Metrics Across Wheat  Organs}
    \label{fig:radar_wheat}
\end{figure}

The results of the model on the MMW are shown in Table \ref{tab:wheat_metrics_percent}, and Figure \ref{fig:radar_wheat} is the visualization of the metrics. The results show that the IoU of panicle, leaf, and stem is 92.82\%, 86.47\%, and 89.76\%, respectively. The segmentation result of wheat panicle is the best, indicating that the model has a good segmentation effect on large targets with less occlusion. The segmentation effect of leaves is slightly worse than that of panicles and stems, which may be due to the irregular edges of leaves or their adhesion to stems, leading to boundary prediction errors. The results for stems indicate that the model can accurately capture the structure and boundaries of stems. Overall, the model can solve the task of high-precision wheat organ segmentation. Compared with the results of rice leaves and stems, it can be seen that the model achieves better results for wheat due to its relatively clear organ boundaries.
\begin{table}[htbp]
    \centering
    \caption{Evaluation Metrics for Different Wheat Organs (Percentage)}
    \label{tab:wheat_metrics_percent}
    \setlength{\tabcolsep}{3pt} 
    \begin{tabular}{lcccc}
    \toprule
    Organ   & IoU    & Precision  & Recall   & F1-score \\
    \midrule
    Panicle   & 92.82      & 99.40         & 93.34       & 96.24        \\
    Leaf    & 86.47      & 98.04         & 88.27       & 92.61        \\
    Stem    & 89.76      &  96.12        & 93.23      & 94.45        \\
    \bottomrule
\end{tabular}
\end{table}

Tables \ref{tab:compare} and \ref{tab:wheat compare} present the performance comparison results of mainstream deep learning-based instance segmentation algorithms. Specifically, The unsupervised method CrossPoint\cite{Afham_2022_CVPR} is a representative algorithm for current unsupervised instance segmentation. The interactive method Agile3D\cite{yue2023agile3d} is the latest supervised SOTA model, where the number after the "@" symbol indicates the number of user clicks required for 3D point cloud segmentation (the accuracy of the model usually increases with the number of interactions). To maintain consistency in comparison benchmarks, this paper only compares its first interaction results. The fully supervised method oneformer3D\cite{kolodiazhnyi2024oneformer3d} is currently the best fully supervised model, and its results can be regarded as the ideal upper limit for this task. This hierarchical comparison architecture clearly shows the performance differences of algorithms under different supervision paradigms.

\begin{table}[h]
    \centering
    \caption{Comparison of Mainstream Segmentation Methods on the MMR Dataset}
    \label{tab:compare}
    \small
    \begin{tabular}{lcccc}
        \toprule
        \textbf{Method} & \multicolumn{3}{c}{\textbf{IoU Per Category (\%)}} & \textbf{mIoU} \\
        \cmidrule(lr){2-4}
        & \textbf{Grain} & \textbf{Leaf} & \textbf{Stem} &  \\
        \midrule
        CrossPoint & 39.64  & 18.13 &  12.46& 23.41 \\
        Agile3D @1 & 69.60 & 70.60 & 27.25 & 55.82 \\
        oneformer3D &79.81  & 77.55 & 77.18 & 78.18 \\
        \textbf{IPENS} & \textbf{61.48} & \textbf{69.54} & \textbf{60.13} & \textbf{63.72} \\
        \bottomrule
    \end{tabular}
\end{table}
\begin{table}[h]
    \centering
    \caption{Comparison of Mainstream Segmentation Methods on the MMW Dataset(on processing)}
    \label{tab:wheat compare}
    \small
    \begin{tabular}{lcccc}
        \toprule
        \textbf{Method} & \multicolumn{3}{c}{\textbf{IoU Per Category (\%)}} & \textbf{mIoU} \\
        \cmidrule(lr){2-4}
        & \textbf{Panicle} & \textbf{Leaf} & \textbf{Stem} &  \\
        \midrule
        CrossPoint & 27.79 &12.63  & 9.09 & 16.50 \\
        Agile3D @1 & - & - & - & - \\
        oneformer3D & 99.26 & 98.16 &95.15 & 97.52 \\
        \textbf{IPENS} & \textbf{92.82} & \textbf{86.47} & \textbf{89.76 } & \textbf{89.68} \\
        \bottomrule
    \end{tabular}
\end{table}

The results show that oneformer3D performs optimally on both data sets (mIoU: Rice 78.18\%/Wheat 97.52\%), reflecting its theoretical advantage of full supervision with sufficient labeled data. Our method is significantly better than the unsupervised algorithm CrossPoint on both crops (mIoU: Rice 23.41\%/Wheat 16.50\%), demonstrating the effectiveness of IPENS in the absence of labels. Compared with the first interaction results of Agile3D, the grain segmentation effect of IPENS on the rice data set is slightly lower by 8.12\%, and the leaf effect is close. It is noteworthy that the segmentation performance of our method on stems is much higher than that of Agile3D.
This indicates that the unsupervised interactive algorithm IPENS can perform at the mainstream level on rice and wheat data sets, demonstrating the feasibility of applying unlabeled data in agricultural scenarios.

This paper presents the visualization results of target point cloud extraction on multiple species in Section \ref{sec:multi_species_point_cloud_visualization}, showing the potential of the model for subsequent cross-species applications.

\subsubsection{Time performance analysis}

\begin{figure*}[h]  
    \centerline{\includegraphics[width=0.9\linewidth]{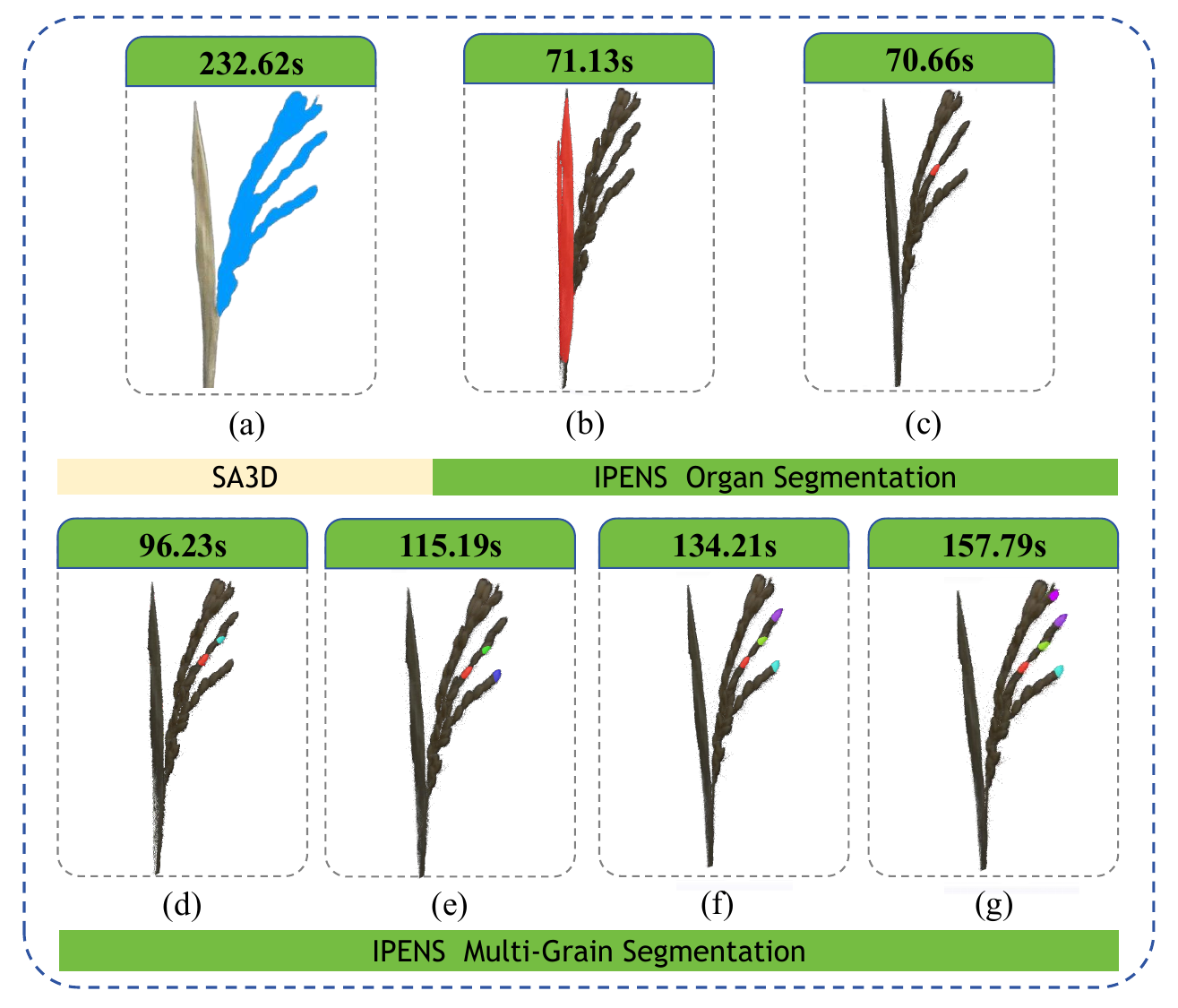}}  
    \caption{Comparison of target inference time, taking rice as an example. (a) SA3D single-target point cloud segmentation time. (b)-(c) IPENS segmentation time for different organs. (d)-(g) IPENS simultaneous segmentation time for 2 to 5 multi-target.}  
    \label{fig:time}  
\end{figure*}
This paper conducts a quantitative analysis of the correlation between model segmentation efficiency and the number of targets, as shown in Figure \ref{fig:time}. The experiment collected the average segmentation time of SA3D for single targets and IPENS for single and multiple targets.
The experimental results indicate that: (1) Thanks to the customization of light parameters and the application of parallel algorithms, as compared in (a) and (b), IPENS has about a 3.3x acceleration over SA3D. (2) For single-organ segmentation tasks, such as the segmentation time for leaves in (b) and individual grains in (c) being 71.13s and 70.66s respectively, with a time difference of only 0.66\%, indicating that the computational efficiency of the model is less affected by the morphological characteristics of the target organs (i.e., the area of the mask region); (3) When dealing with multi-target segmentation, as shown in (b)-(f), the inference time is positively correlated with the number of targets. This phenomenon is closely related to the computational cost of the multi-mask inverse mapping matrix.
This finding provides a relatively effective solution for synchronous and efficient segmentation of multiple organs in breeding scenarios, which can effectively improve the timeliness of multi-target segmentation and thus shorten the overall cycle of phenotypic analysis in genomic selection breeding.

Table~\ref{tab:pipeline_time} presents the time consumption of the IPENS workflow in Reconstructed 3D Point Cloud Extraction, Point Cloud Export, and Trait Extraction, with a total time of only 3 minutes. This reflects the high efficiency of the module design.

\begin{table}[htbp]
    \centering
    \caption{Time Consumption After Reconstruction in the IPENS Workflow}
    \label{tab:pipeline_time}
    \begin{tabular}{lc}
    \hline
    \textbf{Step} & \textbf{Time} \\ 
    \hline
    3D Point Cloud Extraction & 2\,min \\ 
    Point Cloud Export & 0.5\,min \\ 
    Trait Extraction & 0.5\,min \\ 
    \hline
    \end{tabular}
\end{table}

This paper also discusses in Section \ref{sec:recon_time_consume} the selection of reconstruction algorithms and their corresponding time consumption based on different application scenarios.

\subsection{Phenotypic analysis}
\subsubsection{Analysis of rice grain volume}

\begin{figure}[h]
    \centerline{\includegraphics[width=\linewidth]{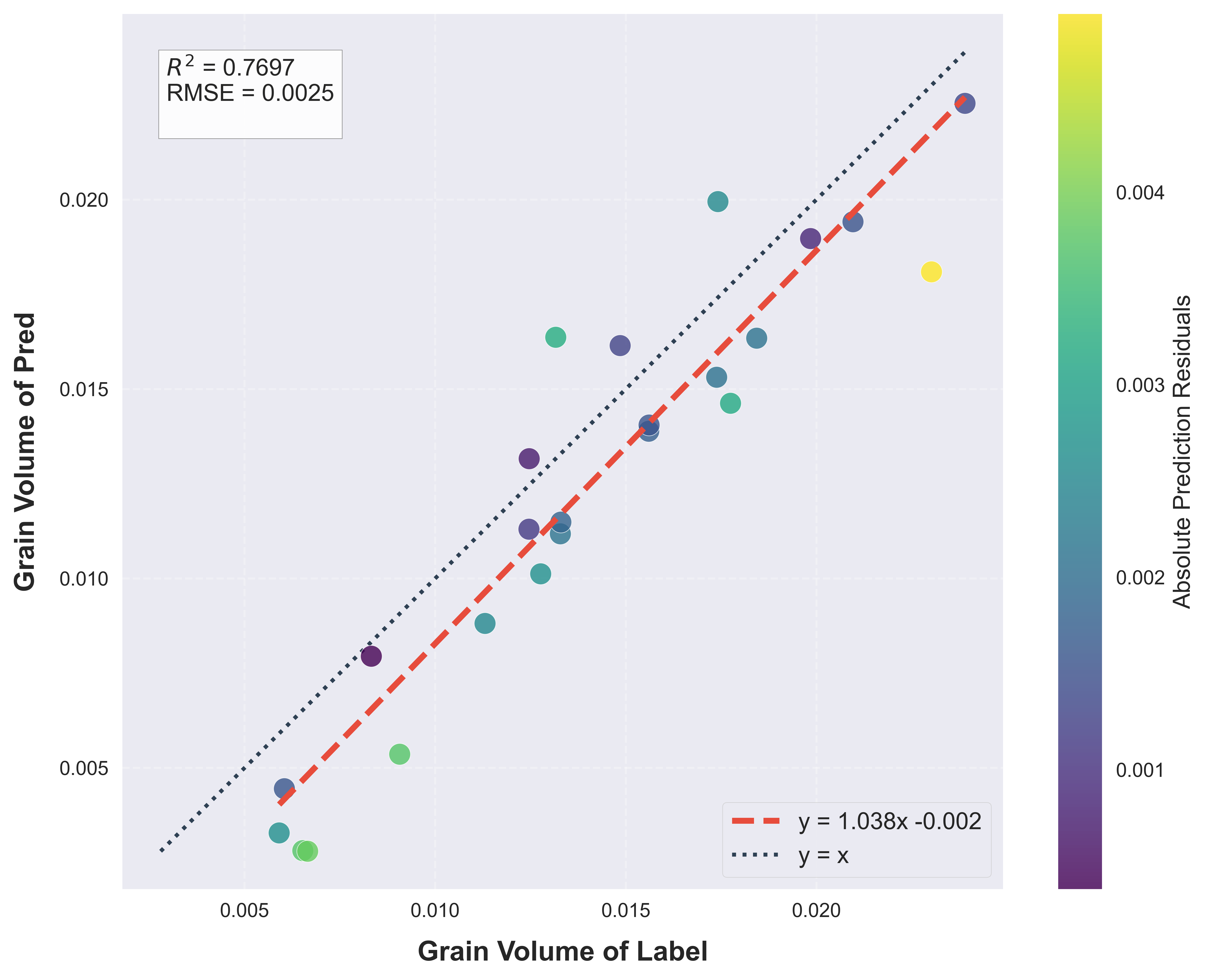}}
    \caption{Correlation between labeled and predicted grain volume extracted by model}
    \label{fig:grain volume}
\end{figure}

Based on the volume calculation algorithm, Figure \ref{fig:grain volume} shows the correlation between the labeled and predicted volume of rice grains, with an R² of 0.7697 and an RMSE of 0.0025.
These results validate the effectiveness of IPENS in extracting grain volume and the accuracy of the 3D point cloud.

\subsubsection{Analysis of wheat panicle volume}
\begin{figure}[h]
    \centerline{\includegraphics[width=\linewidth]{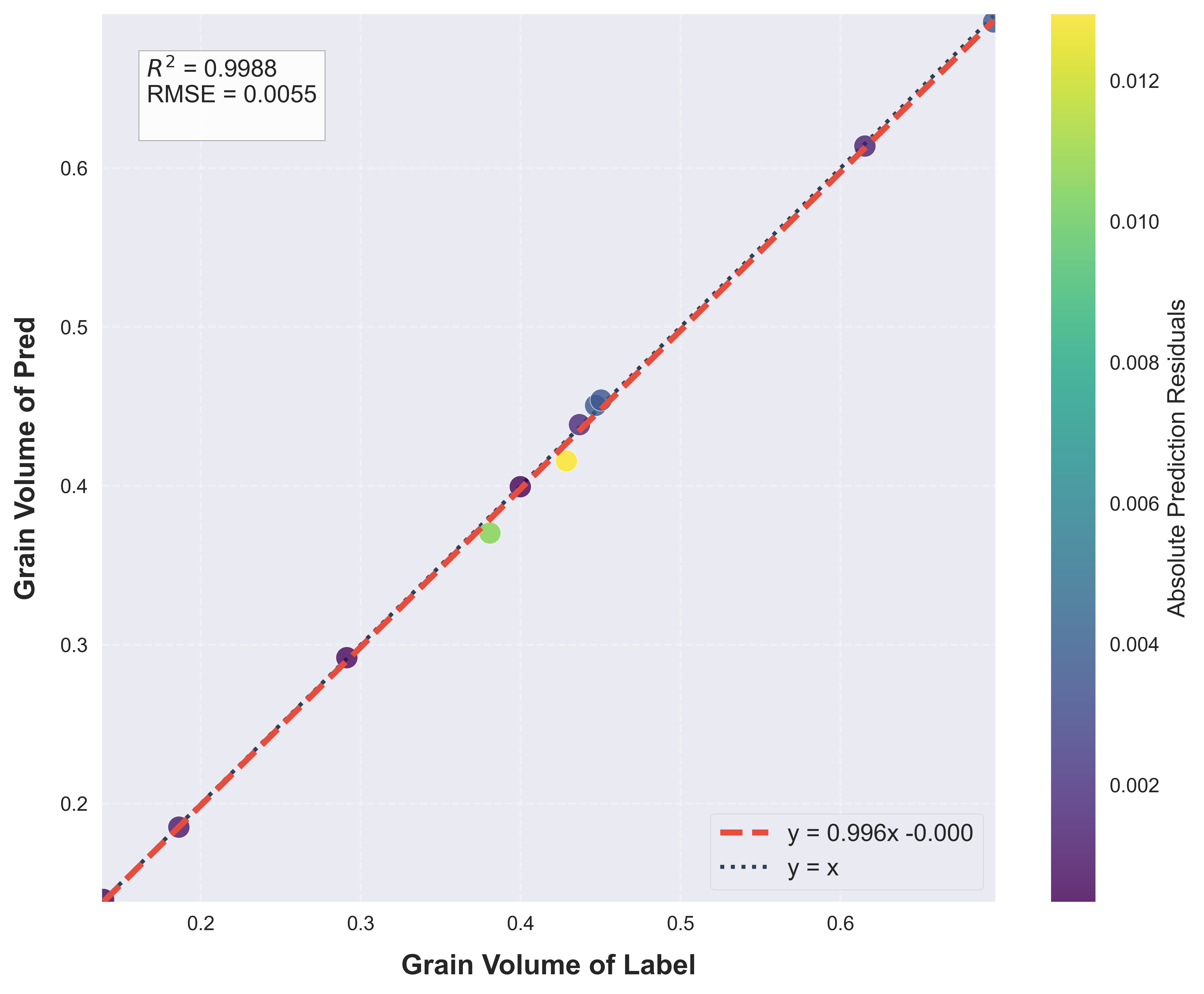}}
    \caption{Correlation between labeled and predicted panicle volume extracted by model}
    \label{fig:panicle volume}
\end{figure}
Figure \ref{fig:panicle volume} shows the correlation between the labeled and predicted volume of wheat panicles, with an R² of 0.9956 and an RMSE of 0.0055. The fitting results nearly overlap with the reference fitting, indicating that the proposed method can achieve high-quality segmentation for wheat panicles.

\subsubsection{Leaf phenotypic analysis}


\begin{table*}
    \caption{Comparison results of rice leaf surface area (in cm²)}
    \centering
    \label{tab:area_results}
    \begin{tabular}{lcccccc}
    \toprule
    Step  & Label Mean & Pred Mean & RMSE & MAE & R² \\
    \midrule
    Cloud Convex Hull Area  & 98.54 & 109.49 & 28.27 & 18.24 & 0.68 \\
    Repaired Mesh Area  & 79.47 & 89.29 & 19.37 & 13.58 & 0.84 \\
    Subdivision Mesh Area  & 77.33 & 86.80 & 18.93 & 13.21 & 0.84 \\
    \bottomrule
    \end{tabular}
    \end{table*}



\begin{table}[h]
    \centering
    \caption{Statistical analysis of rice leaf length and width (in cm)}
    \label{tab:len_wid_results}
\begin{tabular}{lrr}
    \hline
     Metric   &   Length &   Width \\
    \hline
    Label Mean &   24.77 &   2.57 \\
    Pred Mean  &   24.32 &   2.38 \\
    RMSE     &          1.49 &         0.21\\
     MAE      &        1.03 &         0.19  \\
     R²       &        0.97 &       0.87 \\
    \hline
    \end{tabular}
    \end{table}

\begin{table*}
\caption{Comparison results of wheat leaf surface area (in cm²)}
\centering
\label{tab:wheat_scaled_metrics}
\begin{tabular}{lcccccc}
\toprule
Step  & Label Mean & Pred Mean & RMSE & MAE & R² \\
\midrule
Cloud Convex Hull Area  & 22.98 & 23.55 & 1.72 & 1.13 & 0.99 \\
Repaired Mesh Area  & 16.32 & 16.42 & 0.78 & 0.63 & 1.00 \\
Subdivision Mesh Area  & 15.87 & 15.91 & 0.67 & 0.53 & 1.00 \\
\bottomrule
\end{tabular}
\end{table*}

\begin{table}[h]
\centering
\caption{Statistical analysis of wheat leaf length and width (in cm)}
\label{tab:wheat_len_wid_results}
\begin{tabular}{lrr}
\hline
 Metric   &   Length &   Width \\
\hline
Label Mean &   10.23 &   1.00 \\
Pred Mean  &   10.08 &   0.98 \\
RMSE     &          0.23 &         0.15 \\
MAE      &          0.18 &         0.12 \\
R²       &          0.99 &         0.92 \\
\hline
\end{tabular}
\end{table}

Table \ref{tab:area_results} shows the surface area of rice leaves obtained at three different stages of point cloud processing, and compares the results of Label and Pred. 
After preprocessing, the calculation of the convex hull area of the point cloud has a high RMSE (28.27) and MAE (18.24), and the fitting situation R² is 0.68. This indicates that there is a large error in calculating the leaf area by convex hull, which is greatly affected by point cloud noise. 
The prediction error of the area after meshing the leaves has decreased, indicating that the repair treatment may have removed some noise and local abnormalities, improving the result stability. 
After subdividing the mesh, the RMSE and MAE are 18.93 and 13.21, and R² has increased to 0.84. The error indicators at this stage have decreased compared with the previous two steps, and the subdivision algorithm further smooths the surface details, making the overall area estimation more accurate. 
Table \ref{tab:wheat_scaled_metrics} shows the surface area results of wheat leaves. It can be found that after subdivision, the RMSE and MAE reach 0.67 and 0.53, and R² reaches 1.00, indicating the excellent ability of the model in estimating the area of wheat leaves.

Overall, the gradual decrease in error at each stage indicates that the estimation of leaf surface area is gradually stabilizing and demonstrating higher accuracy.

Table \ref{tab:len_wid_results} presents the statistical results of the length and width of rice leaves. It indicates that the length and width measurement method exhibits high accuracy in estimating leaf size. Specifically, the predicted mean value for leaf length (24.32cm) is close to the truth (24.77cm), and the mean width value (2.38 cm) is also slightly lower than the truth (2.57cm), indicating that the calculation method difference is at the millimeter level. Error analysis shows that the RMSE of length and width are 1.49 and 0.21, the MAE are 1.03 and 0.19, and R² reaches 0.97 and 0.87, indicating that the deviation between the predicted value and the truth is small, which verifies the stability of the method in rice leaf estimation. 
Table \ref{tab:wheat_len_wid_results} shows the statistical results of the length and width of wheat leaves. The RMSE of leaf length and width are 0.23 and 0.15, the MAE are 0.18 and 0.12 and the R² reaches 0.99 and 0.92. The deviation is very small and the size difference is at the millimeter level. Combined with Table \ref{tab:wheat compare}, the leaf IoU indicates that a better segmentation result corresponds to excellent phenotypic results, which verifies the effectiveness of the method.

\section{Discussion}
\subsection{Effectiveness of the Proposed Method}
To better understand the working mechanism of IPENS, we provide more discussions. Firstly, the good construction of the NeRF field can derive high-quality point clouds, which is beneficial to the improvement of segmentation quality in subsequent processes. SAM2 is pre-trained on SA-V, the largest general object video segmentation dataset so far. This ensures that SAM2 performs well in various tasks and visual fields. 
Since the point cloud extraction capability of the model depends on the 2D segmentation and propagation capabilities of SAM2, more prompt information will optimize the accuracy of 2D target segmentation, and appropriate post-processing operations to fill holes and smooth edges will make the 3D mask more accurate.
\subsection{Limitation and future prospects}
Due to the limitations of machine video memory and mask storage design, IPENS will encounter insufficient memory problems as the number of targets being segmented simultaneously increases. This has become an obstacle to clicking on a target type such as rice grains and simultaneously segmenting all the grains. This requires gradual optimization of video memory usage design in subsequent work to enhance multi-target segmentation capabilities. 
Secondly, the reconstruction process of IPENS requires data acquisition when the object is in a wind-free condition, which limits its application in real field scenarios. 
To solve this problem, we are developing a field phenotyping unmanned vehicle that uses multiple cameras to expose simultaneously, achieving high-speed multi-view imaging of crops.

\section{Conclusion}
This paper proposes an IPENS framework that combines the 2D segmentation and propagation capabilities of SAM2 with radiance field information to extract 3D target point clouds. A multi-target collaborative point cloud extraction scheme is designed. Two post-processing methods are proposed to optimize the 3D target point clouds based on the SAM2 segmentation results. In an unsupervised interactive manner, the IPENS method achieves grain-level segmentation on a rice dataset, with an mIoU of 63.72\% for grains, leaves, and stems. The R² for grain volume is 0.7697, and the RMSE is 0.0025; the R² for leaf surface area is 0.84, and the RMSE is 18.93; the R² for leaf length and width are 0.97 and 0.87 respectively, with RMSE of 1.49 and 0.21. 
On the wheat dataset, the mIoU for panicles, leaves, and stems is 89.68\%, the R² for panicle volume is 0.9956, and the RMSE is 0.0055; the R² for leaf surface area is 1.0, and the RMSE is 0.67; the R² for leaf length and width are 0.99 and 0.92 respectively, with RMSEs of 0.23 and 0.15. These results demonstrate the effectiveness of the IPENS model in phenotypic extraction of rice and wheat.

The IPENS method achieves unsupervised, interactive, and non-invasive extraction of grain-level point clouds. This can provide high-quality phenotypic data for establishing genotype-phenotype association models, and its efficient extraction process has great potential in promoting large-scale intelligent breeding.

\appendix

\section{Multi-Species Point Cloud Extraction Visualization}\label{sec:multi_species_point_cloud_visualization}
\begin{figure*}[h]
    \centerline{\includegraphics[width=\linewidth]{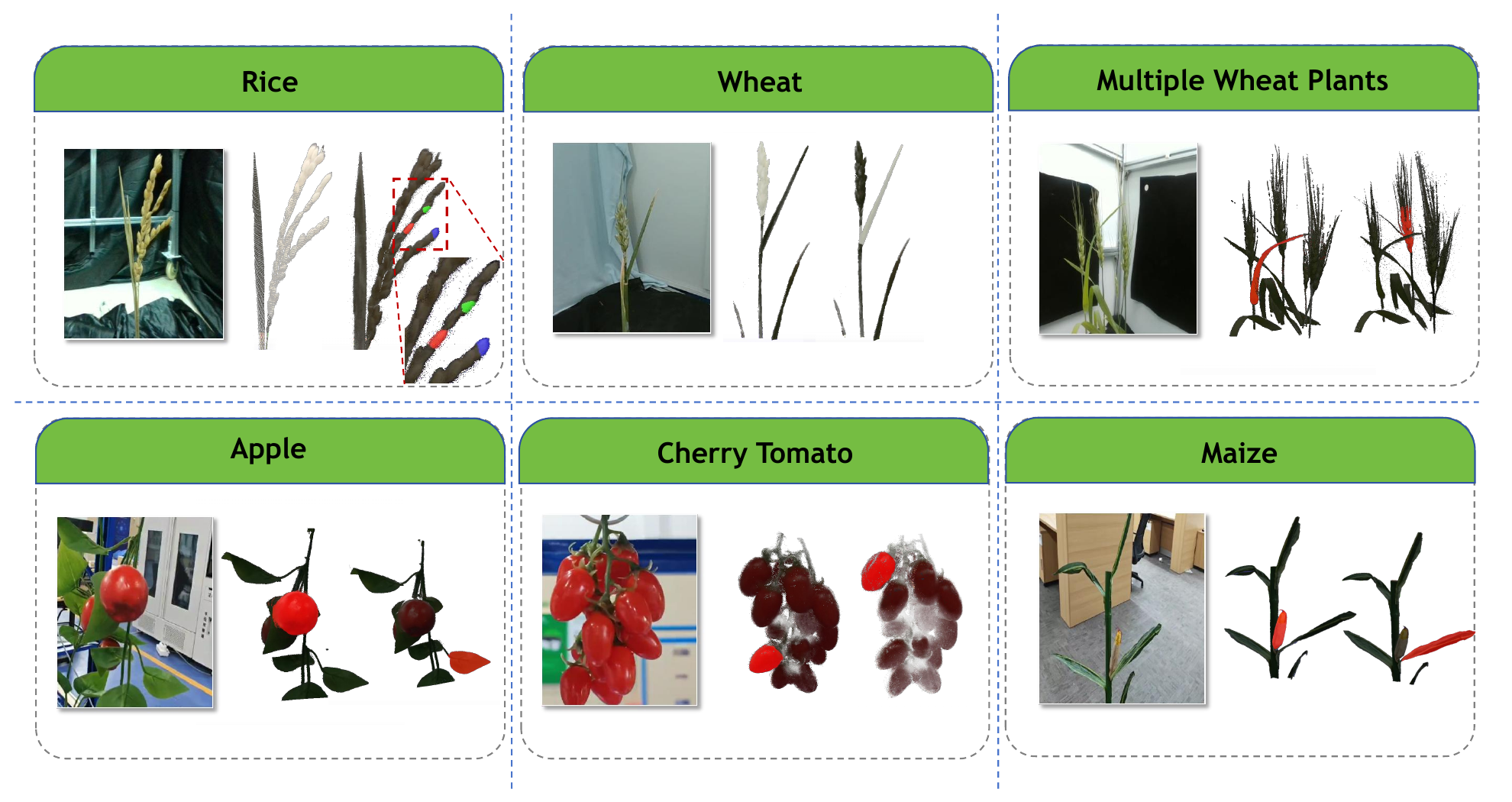}}
    \caption{Visualization results under different crops}
    \label{fig:dingxin}
\end{figure*}
This section presents the visual results of instance segmentation on multiple species. 
As shown in Fig.~\ref{fig:dingxin}, IPENS achieves good segmentation effects on the organs of different crops. It is worth noting that the model has not undergone specialized semantic segmentation training but can still adapt to various scenarios and species, showing strong generalization ability. This verifies the possibility of the model in extracting point clouds across species and varieties in the future.

\section{Time consumption of 3D reconstruction models}\label{sec:recon_time_consume}

Depending on different application scenarios, it is necessary to select an appropriate 3D reconstruction model. Table \ref{tab:model_metrics} shows the reconstruction performance and effects of Instant-NGP, Instant-NGP-bounded, Nerfacto, Nerfacto-big, and Mip-NeRF360. The performance of different 3D reconstruction methods is compared in terms of Training time, Peak Signal to Noise Ratio (PSNR), structural similarity index (SSIM), Learned Perceptual Image Patch Similarity (LPIPS), and Frames Per Second (FPS).


Based on the time cost and performance metrics in the table, the models can be categorized into three groups: 
\begin{itemize}  
    \item Fast Models (<10 minutes): This includes Instant-NGP, Instant-NGP-bounded, and Nerfacto. These models are suitable for real-time monitoring, quick previews, or applications on mobile devices and edge computing. For example, in drone-based field inspections, they can instantly generate 3D models of crops to assess growth conditions; they are also ideal for processing data quickly on resource-constrained devices such as agricultural robots or handheld terminals.  

    \item Medium-Speed Models (~1 hour): Nerfacto-big falls into this category. It is suitable for high-precision agricultural analysis, where a balance between speed and accuracy is required. For instance, it can be used for dynamic monitoring of crop growth in greenhouses, generating high-precision models weekly to track development trends.  

    \item Long-Time Models (Several hours): Models like Mip-NeRF360 belong to this group. They are suitable for reconstructing extremely complex scenes or scenarios with high-fidelity requirements. For example, in the holographic modeling of ancient or famous trees for cultural heritage preservation, or for publishing research papers or generating standard datasets, these models maximize reconstruction quality despite the longer processing time.  

\end{itemize}

\begin{table}[ht]
    \centering
    \caption{Training time, PSNR, SSIM, LPIPS and FPS for Instant-NGP,Instant-NGP-bounded,Nerfacto,Nerfacto-big, Mip-NeRF360  3D reconstruction.}
    \label{tab:model_metrics}
    \setlength{\tabcolsep}{2.6pt} 
    \resizebox{0.48\textwidth}{!}{
    \begin{tabular}{lcccccc}
    \toprule
    Model               & Time     & PSNR   & SSIM  & LPIPS & FPS    \\
    \midrule
    Instant-NGP         & 5 min    & 17.43  & 0.52  & 0.66   & 0.18  \\
    Instant-NGP-bounded & 9 min    & 27.98  & 0.81  & 0.37   & 1.08   \\
    Nerfacto            & 8 min    & 23.96  & 0.76  & 0.36   & 1.96   \\
    Nerfacto-big        & 55 min   & 24.19  & 0.75  & 0.34   & 0.66      \\
    Mip-NeRF360         & 30 h     & 29.63  & 0.83  & 0.32   & 0.05      \\
    
    \bottomrule
\end{tabular}
    }
\end{table}

\bibliography{mybibfile}

\end{document}